\documentclass[letterpaper]{article} 
\usepackage{aaai2026}  
\usepackage{times}  
\usepackage{helvet}  
\usepackage{courier}  
\usepackage[hyphens]{url}  
\usepackage{graphicx} 
\usepackage{natbib}  
\usepackage{caption} 
\usepackage{subcaption}
\usepackage[fleqn]{amsmath}
\usepackage{amssymb}
\usepackage{algorithm}
\newcommand{\ZZ}{\mathbb{Z}}
\newcommand{\TTT}{\mathcal{T}}
\usepackage{stmaryrd}
\usepackage{listings}
\DeclareCaptionStyle{ruled}{labelfont=normalfont,labelsep=colon,strut=off} 
\floatstyle{ruled}
\newfloat{listing}{tb}{lst}{}
\floatname{listing}{Listing}
\usepackage{makecell}
\usepackage{algorithm}
\usepackage[noend]{algpseudocode}

\usepackage{pgfplots}
\usepgfplotslibrary{groupplots} 

\usepackage{enumitem}

\usepackage{bm}
\usepackage{multirow}
\usepackage{booktabs}
\usepackage{arydshln} 

\usepackage{tikz}
\usepackage{tikzpeople}
\usetikzlibrary{decorations.pathreplacing,calligraphy}
\usetikzlibrary{positioning,shadows,arrows,shapes.arrows,shapes.symbols,3d,calc,shapes}
\usepackage{pgfplots} 
\usetikzlibrary{positioning,fit}

\tikzset{
  module/.style={rectangle, draw, rounded corners, align=center, minimum width=2.6cm, minimum height=0.8cm},
  attn/.style={rectangle, draw, rounded corners, fill=purple!20, align=center, minimum width=2.6cm, minimum height=0.8cm},
  soft/.style={rectangle, draw, rounded corners, fill=green!20, align=center, minimum width=2.6cm, minimum height=0.8cm},
  norm/.style={rectangle, draw, rounded corners, fill=orange!30, align=center, minimum width=2.6cm, minimum height=0.8cm},
  class/.style={rectangle, draw, rounded corners, fill=red!30, align=center, minimum width=2.6cm, minimum height=0.8cm}
}
\title{Privacy-Preserving Inference for Quantized BERT Models}

\begin{document}

\author{
    Tianpei Lu\textsuperscript{\rm 1,}\textsuperscript{\rm 3,},
    Bingsheng Zhang\textsuperscript{\rm 1,}\textsuperscript{\rm 3,}\thanks{Bingsheng Zhang is the corresponding author.},
    Lekun Peng\textsuperscript{\rm 1},
    Bowen Zheng\textsuperscript{\rm 1},\\
    Lichun Li\textsuperscript{\rm 2},
    Kui Ren\textsuperscript{\rm 1}
}
\affiliations{
    \textsuperscript{\rm 1}The State Key Laboratory of Blockchain and Data Security, Zhejiang University, China,
    \textsuperscript{\rm 2}Ant Group, China, \\
    \textsuperscript{\rm 3}Hangzhou High-Tech Zone (Binjiang) Institute of Blockchain and Data Security,\\
\{lutianpei, bingsheng\}@zju.edu.cn

%

}
\maketitle

\begin{abstract}
With the increasing deployment of generative machine learning models in privacy-sensitive domains such as healthcare and personalized services, ensuring secure inference has become a critical challenge. Secure multi-party computation (MPC) enables privacy-preserving model inference but suffers from high communication and computation overhead. The main bottleneck lies in the expensive secure evaluation of floating-point operations.
Quantization offers a promising solution by converting floating-point operations into lower-precision integer computations, significantly reducing overhead. However, existing MPC-based quantized inference methods either rely on public quantization parameters—posing privacy risks—or suffer from inefficiencies, particularly in handling nonlinear functions such as activations and softmax. In this work, we propose a fine-grained, layer-wise quantization scheme and  support 1-bit weight fully connected layers in a secure setting. We design a multi-input lookup table protocol to evaluate softmax efficiently and securely. Furthermore, we use dual secret sharing schemes and perform precision conversions via lookup tables, eliminating truncation overhead entirely. Experimental evaluation on BERT-base models demonstrates that our approach achieves up to $8\times$ speedup compared to Lu \emph{et al}. (NDSS 25), $9\times$ speedup compared to Gupta \emph{et al}. (PETS 24) and $22 \times$ speedup compared to Knott \emph{et al}. (NeurIPS 21).

\end{abstract}

%


\section{Introduction}\label{sec:introduction}

With the widespread deployment of machine learning models in privacy-sensitive domains such as healthcare and personalized services, ensuring data confidentiality has emerged as a paramount concern. Privacy-preserving machine learning (PPML) techniques enable model inference while maintaining confidentiality, with secure multi-party computation (MPC) being the most popular approach. State-of-the-art (SOTA) PPML platforms, including Orca~\cite{orca}, Falcon~\cite{WTB+-21}, ABY$^3$~\cite{MR18}, and SecureML~\cite{MZ17}, predominantly leverage MPC frameworks. 
However, PPML techniques incur substantial communication overhead, especially for large-scale language models. For instance, BERT-base with 12 transformer layers and over 110 million parameters requires billions of secure operations for a single inference. MPC-based inference can generate several gigabytes of communication, creating barriers for real-world deployment.
  

One of the key challenges in privacy-preserving machine learning is that most machine learning models are designed to operate using floating-point arithmetic, which incurs prohibitive overhead when implemented under MPC. The most efficient approach is to approximate these computations using fixed-point arithmetic. In fixed-point representation, a real number $x$ is encoded as an integer $\lfloor x \cdot 2^\ell \rfloor$ before being processed under MPC, where $\ell$ denotes the number of fractional bits used to represent precision. However, after performing a multiplication $x \cdot y$, the result becomes $x \cdot y \cdot 2^{2\ell}$, and thus a truncation protocol is required to reduce it back to $x \cdot y \cdot 2^\ell$. This truncation step is costly. In current implementations, probabilistic truncation~\cite{previous1,MZ17,MR18,RRK+-20,HLH+-22} is widely adopted, which necessitates reserving a large number of high-order bits as a tolerance margin to prevent truncation errors\footnote{The truncation error is caused by the wrap bit. For example, in secret sharing, $x_1 + x_2 \bmod 2^\ell = x$. If a wrap occurs, then $x_1 + x_2 = x + 2^\ell$. When truncating by $k$ bits, we compute $
{x_1}/{2^k} + {x_2}/{2^k} \bmod 2^\ell
= x/{2^k} + 2^{\ell - k}$, which introduces an error of $2^{\ell - k}$.
}. Specifically, for a truncation error bound of $2^{-k}$, $k$ extra bits are needed; the larger $k$ is, the lower the probability of error.
For multiplication of inputs with $\ell_x$-bit precision, at least $2\ell_x$ bits are required to avoid overflow. When accounting for error tolerance in truncation, the overall bit-width expands to $2\ell_x + k$. For example, when $\ell_x = 16$ and $k = 32$, the computation must be performed over a 64-bit ring. Moreover, in scenarios such as matrix summation, where intermediate results may exceed $2\ell_x$, an even larger bit-width may be necessary to ensure correctness and prevent overflow.

Quantization has emerged as a promising technique to enhance the performance of privacy-preserving machine learning (PPML)~\cite{comment1,comment3,comment4,comment5}, particularly in distributed settings such as federated learning. However, its application in secure multiparty computation (MPC)-based PPML remains at an early stage. Several works, including those by Riazi \textit{et al.}\cite{xonn}, Agrawal \textit{et al.}\cite{mpcqu2}, and Keller \textit{et al.}~\cite{mpcqu3}, adopt traditional MPC protocols to evaluate quantized models. To optimize performance, these studies often rely on public, fixed quantization parameters—typically powers of two—which, while computationally efficient, can degrade model accuracy and risk leaking sensitive model information.
Lu~\textit{et al.}~\cite{our1} were the first to solve the truncation problem in quantized models through table lookup evaluation.  While their approach ensures highly efficient online phase computation, it uses lookup tables to handle multiplication. This design results in a total of 256 bits of communication per multiplication gate (from 8-bit input to table lookup) in the offline phase, which makes the scheme challenging to deploy in practical scenarios.
In this work, we address the limitations of Lu et al.’s method. Instead of using lookup tables for multiplication, which we find inefficient for matrix multiplication compared to the 2-out-of-3 protocol, we use lookup tables solely to handle truncation. Specifically, we truncate by discarding the lower bits (e.g., keeping the upper 4 bits from 16 bits) and then use a lookup table to expand the 4-bit value back to 16 bits. This avoids high-bit errors caused by truncation\footnote{When truncating while also reducing the modulus to $2^{\ell-k}$, no truncation error occurs.
If a wrap happens so that $x_1+x_2=x+2^\ell$, then ${x_1}/{2^k}+{x_2}/{2^k}\bmod 2^{\ell-k}=x$, eliminating the error term.
}~\cite{previous1}. In addition, by designing a multi-input lookup table protocol, we enable the efficient evaluation of functions such as Softmax.
Overall, the contributions of this paper can be summarized as follows:

\begin{itemize}
\item We analyze the impact of weight and activation precision on model performance, and train the model with 1-bit weights and 4-bit activations. Under this quantization setting, privacy-preserving inference can be efficiently supported through secure multiplication combined with lookup table protocols.
\item We design a multi-input lookup table protocol and, based on it, develop an efficient and systematic softmax evaluation protocol tailored for quantized models.
\item We employ two types of secret sharing schemes to achieve layer-specific optimal implementations. To bridge different precisions across these schemes, we use lookup table evaluation to perform share conversion between different precisions, eliminating the overhead associated with truncation. As a result, our approach achieves up to $8\times$ speedup compared to Lu \emph{et al}. (NDSS 25), $9\times$ speedup compared to Gupta \emph{et al}. (PETS 24) and $22 \times$ speedup compared to Knott \emph{et al}. (NeurIPS 21). 
\end{itemize}
%

%
%

\section{Preliminaries} \label{sec:preliminary}
%


\noindent \textbf{Notations.}  \label{sub:notation} Let $\mathcal{P}:=\{P_0,P_1,P_2\}$ be the three MPC parties. 
We use $:=$ to denote the definition. 
We use $\mathbb{Z}_{2^\ell}$ to denote the set $\{0, \ldots, 2^\ell - 1\}$. For a set $S$, $S^\ell$ indicates the $\ell$ dimension vector space over the set $S$, namely, 
$S^\ell := \{ (s_0, s_2, \ldots, s_{\ell-1}) \mid s_i \in S \}$. 
We encode value $x \in [-2^{\ell-1}, 2^{\ell-1})$ to the ring $\ZZ_{2^\ell}$.
Here, for $x \in [0, 2^\ell)$, positive values remain unchanged, while negative values $x$ are encoded as $2^\ell + x \in [2^{\ell - 1}, 2^\ell)$.
 $\ell$ is the bit length of an element in $\mathbb{Z}_{2^\ell}$. $x||y$ refers to the concatenation of $x$ and $y$. 
We use $\mathsf{trc}(x, k)$ to denote the truncation of $x$ to its first $k$ bits.

\smallskip
\noindent \textbf{Two Party Additive Secret Sharing.} We denote two party additive secret sharing over ring $\ZZ_{2^\ell}$ as  $\llbracket x \rrbracket^\ell:=(\llbracket x \rrbracket^\ell_1, \llbracket x \rrbracket^\ell_2)$. It holds that $\llbracket x \rrbracket^\ell_1 + \llbracket x \rrbracket^\ell_2 \mod 2^{\ell}= x$. The party $P_{1}$ holds $\llbracket x \rrbracket^\ell_1$ and  $P_{2}$ holds $\llbracket x \rrbracket^\ell_2$. It possesses additive homomorphic properties, meaning that $\llbracket x + y \rrbracket^\ell = (\llbracket x \rrbracket^\ell_1 + \llbracket y \rrbracket^\ell_1 \mod 2^{\ell}, \llbracket x \rrbracket^\ell_2  + \llbracket y \rrbracket^\ell_2\mod 2^{\ell})$. 
When party $P$ wants to secret-share a value $x$, it first agrees with party $P_1$ on the same seed $s$. Then, both $P$ and $P_1$ use $s$ to generate the random share $[x]_1$. Party $P$ computes $[x]_2 = x - [x]_1$, and sends $[x]_2$ to party $P_2$. We use $\Pi_\mathsf{share}(x, P)$ to denote the secret share scheme. 
$P_1$ and $P_2$ can exchange $[x]_1$ and $[x]_2$ with each other to reveal the value $x = [x]_1 + [x]_2$.

\smallskip
\noindent \textbf{Secure Three Party Replicated Secret Sharing.} Replicated Secret Sharing (RSS) is a secret sharing scheme commonly used in secure multi-party computation. It requires at least three parties. Compared to two-party computation, its advantage lies in not relying on public-key cryptography, enabling more efficient secure computation. In our setting, we define the 3PC-RSS over ring $\ZZ_{2^\ell}$ as $\langle x\rangle^\ell := (\langle x\rangle^\ell_0, \langle x\rangle^\ell_1, \langle x\rangle^\ell_2)$. The party $P_{i-1}$ and $P_{i+1}$ both hold $\langle\cdot\rangle_{i}$. It holds that $\langle x \rangle_0 + \langle x \rangle_1 + \langle x\rangle_2 \mod 2^\ell = x$. 
Similarly, replicated secret sharing also possesses homomorphic properties, namely, $\langle x + y \rangle^\ell = (\langle x \rangle^\ell_0 + \langle y \rangle^\ell_0, \langle x \rangle^\ell_1 + \langle y \rangle^\ell_1, \langle x \rangle^\ell_2  + \langle y \rangle^\ell_2)$.
3PC-RSS is well-suited for evaluating multiplication protocols, especially matrix multiplication. This is because the communication cost of replicated secret sharing depends only on the output dimension. When computing vector inner products, since the output is a scalar, the required communication cost is constant. Below, we present the inner product protocol under replicated secret sharing.
For inner product $z = \sum^N_{j=1} x_j\cdot y_j$ with two input vector $\{\langle x_j\rangle\}_{j \in \ZZ_N}$, $\{\langle y_j\rangle\}_{j\in \ZZ_N}$ and output inner product result $\langle z \rangle$, each party $P_i$ perform 
\begin{equation}
\begin{aligned}
	z_i = \sum^{N-1}_{j = 0} \langle x_j\rangle_{i-1}  \cdot \langle y_j\rangle_{i+1} + \langle x_j\rangle_{i+1}  \cdot& \langle y_j\rangle_{i-1} \\ +& \langle x_j\rangle_{i+1}  \cdot \langle y_j\rangle_{i+1}  \enspace . \nonumber
\end{aligned}
\end{equation}
It holds that $z_0 + z_1 + z_2 = z$ is the inner product result. Then $P_i$ send $z_i$ to $P_{i + 1}$ to recover the 3PC-RSS form.

\smallskip
\noindent\textbf{Lookup Table.} The lookup table $\TTT \in \ZZ^{2^{\ell'}}_{2^\ell}$ for function $f:\{0,1\}^{\ell'} \rightarrow \{0,1\}^{\ell}$ traverse all possible inputs of $f$.  It accepts $\ell'$ bits input and output $\ell$ bits message. The lookup table $\TTT$ can be viewed as a vector. The $r^\mathsf{th}$ item of lookup table $\TTT$ stores the result of $f$ with input $r$.  We use $\TTT(r)$ to denote the $r^\mathsf{th}$ entry of the lookup table.

\smallskip
\noindent \textbf{Oblivious Maximum Evaluation.} Oblivious Maximum Evaluation is used to identify the maximum value from a set of secret-shared values while preserving privacy. In our setting, it takes as input a vector of secret shares and outputs the secret share of the maximum value. Asharov et al.~\cite{sortref} proposed a radix sort algorithm for the three-party computation setting, where the number of sorting rounds depends only on the bit-length of the elements being sorted. In our work, we leverage Asharov et al.'s sorting algorithm to perform maximum evaluation: we first sort the input vector in ascending order and then select the last element as the maximum. We denote the above the maximum protocol as $\Pi_{\mathsf{max}}$.

\smallskip
\noindent \textbf{System Architecture and Threat Model.} Our framework involves three parties: the model owner $P_0$, the data owner $P_1$, and the computing assistant $P_2$. In this framework, the model owner publicly reveals the embedding parameters. The data owner first performs the embedding computation locally, and then quantizes the resulting embeddings into 4-bit values. These quantized embeddings are secret-shared among $P_0$, $P_1$, and $P_2$ using RSS. Our BERT inference framework ensures privacy in the semi-honest adversarial setting, meaning that as long as all parties follow the protocol, no party can obtain any information about the model weights or the input tokens.

\begin{figure}[tbp]
  \centering
\begin{tikzpicture}
\begin{semilogxaxis}[  
    width=6.8cm,
    height=5cm,
    xlabel={Activation bits},
    ylabel={Model accuracy},
    xmin=1, xmax=40,
    log basis x=2,
    x dir=reverse, 
    xtick={32,8,4,2,1},
    xticklabels={32,8,4,2,1}, 
    ymajorgrids=true,
    grid style=dashed,
    line width=0.9pt
]

\addplot[
    blue,
    thick,
    mark=*,
] coordinates {
    (32,85)
    (8,84)
    (4,83)
    (2,81)
    (1,75)
};

\end{semilogxaxis}
\end{tikzpicture}
  \caption{Model accuracy under 1-bit weight and different activation bit-widths}\label{fig:accuracy-bits}
\end{figure}
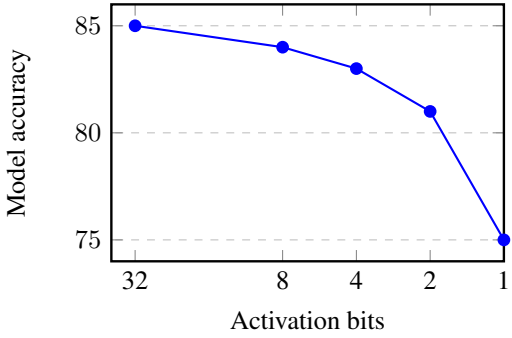
\section{Our BERT Model Structure}\label{sec:bitnet}

The quantization process begins by applying low-bit quantization to the model parameters, including weights and activations (Activations refer to all intermediate variables within the model). Specifically, weights are binarized i.e., to 1 bit, by centering them around zero and applying a sign function, while activations are quantized to integers (e.g., 4-bit, 8-bit) based on their value distribution—either symmetric for values ranging from negative to positive, or asymmetric for strictly non-negative activations. Each quantized layer uses a layer-specific scaling factor $\alpha$ to minimize quantization error, with the option to fine-tune both scaling factors  using gradient-based optimization during training.
After this initial quantization, we perform knowledge distillation to compensate for the loss of accuracy caused by low-bit representations. In this stage, a full-precision teacher model guides the training of the quantized student model. By aligning intermediate representations or output logits, the student learns to approximate the teacher's behavior despite its reduced precision. This two-step procedure—quantization followed by distillation—enables efficient low-bit inference while preserving high model performance.

We conducted experiments under the framework proposed by Liu~\emph{et al}.~\cite{BIT}, evaluating model accuracy across various activation bits. As illustrated in Fig.~\ref{fig:accuracy-bits}, each weights are quantized to $1$ bits and we benchmark different activation bits. Among these configurations, we found that 4-bit activation offers the best trade-off between model performance and efficiency. Based on this observation, we adopt the 4-bit activation setting as the foundation for our overall architecture.



\begin{figure}[tbp]
  \centering\includegraphics[width=\linewidth]{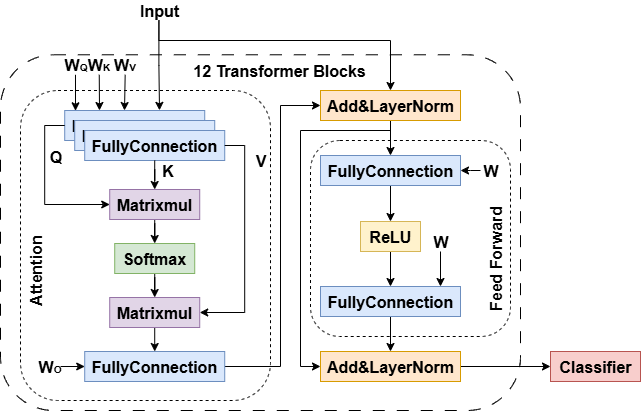}
  \caption{Overview of BitNet}\label{fig:BitNet}
\end{figure}

\noindent \textbf{Model Structure.}\label{sub:evaluation}
Our protocol is divided into a linear computation part and a nonlinear part.
The linear part includes fully connected layers and matrix multiplications.
For the fully connected layers, the parameters come from within the model and are quantized to 1 bit. The quantized weights are represented as ${W} \in {\{-1,1\}}^{m \times n}$, where $m$ is the number of output features and $n$ is the number of input features. The input activations are represented as $x \in {\{-8, -7, \ldots, 6, 7\}}^n$ for signed activations or $x \in {\{0, 1, \ldots, 14, 15\}}^n$ for unsigned activations. The output of the linear layer is computed as follows:
\begin{equation}\label{eq:linear}
  {z}= {W} \cdot x = \alpha_W \cdot \alpha_x \cdot {W} \cdot {x}
\end{equation}
\begin{equation}
  {z'} = \alpha_z \cdot \text{clip}\left(\lfloor \frac{z^i}{\alpha_z}\rfloor, -8, 7\right)
\end{equation}
For the matrix multiplication, we use the same quantization scheme as in the fully connected layer. The only difference is that both inputs are activations, and we quantize both of them to 4 bits. 
For the nonlinear layers, including ReLU, Softmax, and others, these operators do not involve internal parameters; all the inputs are activations, which are directly quantized to 4 bits. 
Take {Softmax} layers as an example, we compute the quantized activations as follows:
\begin{equation}\label{eq:softmax}
  \begin{split}
  z &= \alpha_{z} \cdot \text{clip}(\lfloor \frac{\text{Softmax}(y\cdot \alpha_x)}{\alpha_z}\rfloor, 0, 15) 
  \end{split}
\end{equation}


\section{Our New Technique}\label{sec:quantization}

The core logic of our framework is to convert floating-point computations in transformer models into integer computations through model quantization. We use Replicated Secret Sharing (RSS) to evaluate multiplications, and utilize two-party additive secret sharing held by $P_1$ and $P_2$ to evaluate lookup tables. We first introduce the lookup table evaluation protocol.

\subsection{Secure Lookup Table Evaluation.}
In our designed multi-input lookup table protocol, we first introduce how to securely evaluate a lookup table. In the three-party setting, secure lookup table evaluation can be implemented quite simply~\cite{our1}. Specifically, for a lookup table $\TTT$ held by $P_0$ with $\ell'$ bit input, which means it contains $2^{\ell'}$ items, $P_0$ first randomly generates an offset value $\Delta$, then left-shifts the table $\TTT$ by $\Delta$ positions and secret share the shifted table to $P_1$ and $P_2$, namely, $\llbracket \TTT'\rrbracket := (\llbracket \TTT(\Delta) \rrbracket, \ldots, \llbracket \TTT(\Delta-1) \rrbracket)$. At the same time, $\Delta$ is also secret-shared to $P_1$ and $P_2$.
Then, when $P_1$ and $P_2$ input the secret share $\llbracket x \rrbracket$, they only need to compute $\llbracket x \rrbracket  - \llbracket \Delta \rrbracket$, and open it. After that, $P_1$ and $P_2$ locally take the $(x - \Delta)$-th entry from the secret-shared lookup table as the output. The correctness can be easily verified: $\TTT'(x - \Delta) = \TTT(x - \Delta + \Delta) = \TTT(x)$. Fig.~\ref{alg:look} depicts this algorithm.

\begin{figure}
\includegraphics[scale=0.21]{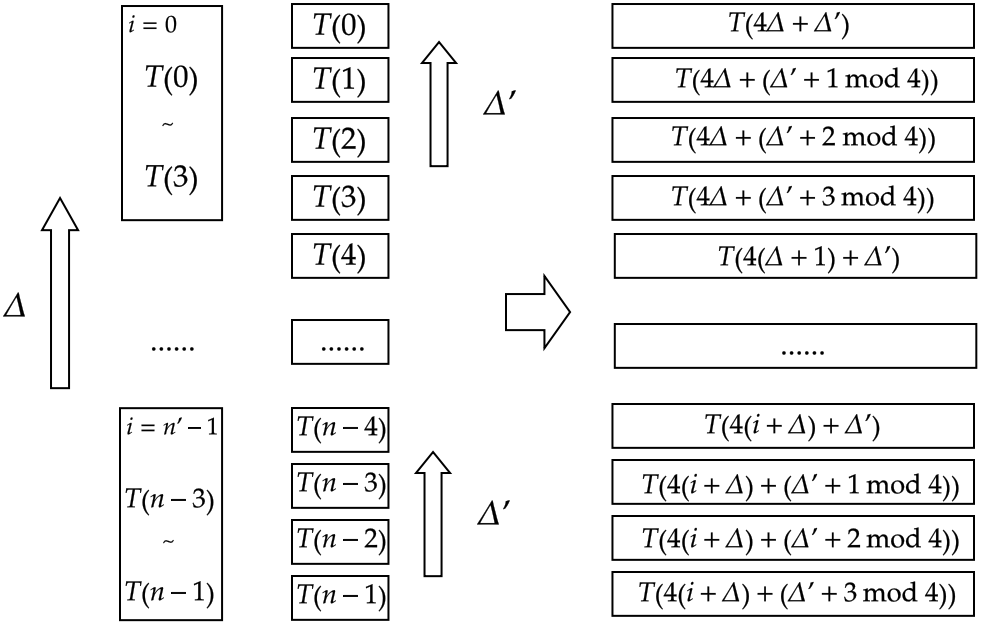}
  \caption{Lookup Table with Separate Inputs. (Take $\ell'=\ell = 2$)}%
  \label{fig:look}%
\end{figure}

\begin{algorithm}[ht]
  \textbf{Input}: lookup table $\mathcal{T} \in \ZZ^{2^{\ell'}}_{2^\ell}$ provided by $P_0$, 2PC additive shared value $\llbracket x \rrbracket^{\ell'}$ held by $P_1$ and $P_2$ \\
  \textbf{Output}: shares of $\mathcal{T}(x)$ 
        \begin{algorithmic}[1]
        \State $P_0$ picks random $\Delta \in \ZZ_{2^{\ell'}}$ and left-shifts $\mathcal{T}$ by $\Delta$ positions;
        \State $P_0$ invoke $\llbracket \mathcal{T} \rrbracket^\ell \leftarrow \Pi_\mathsf{share}(\mathcal{T}, P_0)$ and $\llbracket \Delta \rrbracket^{\ell'} \leftarrow \Pi_\mathsf{share}(\Delta, P_0)$;
        \State $P_1$ and $P_2$ calculate $\llbracket \delta \rrbracket^{\ell'}   = \llbracket x \rrbracket - \llbracket \Delta \rrbracket^{\ell'}  $, reveal $\delta$ to each other;
        \State $P_1$ and $P_2$ set output as $\llbracket \mathcal{T}(\delta) \rrbracket^{\ell} $;
       \end{algorithmic}
    \caption{Lookup table evaluation $\Pi_{\mathsf{look}}$.}
    \label{alg:look}
\end{algorithm}
In this work, we extend the single-input lookup table to support multiple inputs.
Compared to single-input lookup tables, multi-input lookup tables are more complex. Below we intuitively explain the challenges of multi-input settings using a two-input example. For a total input of $\ell'$ bits, it is divided into two parts $x$ and $y$, each of $\ell'/2$ bits, and secret-shared under $\ell'/2$-bit secret sharing. Directly concatenating the two secret shares incurs significant overhead. For example, one may first convert $x$ and $y$ from $\ell'/2$-bit secret shares to $\ell'$-bit secret shares, then compute $x \cdot 2^{\ell'/2} + y$, and finally evaluate a single-input lookup table. However, converting from $\ell'/2$-bit to $\ell'$-bit secret sharing is a nonlinear operation with high cost. We design a multi-input lookup table evaluation scheme whose overhead matches that of the single-input case. Our scheme is illustrated in Fig~\ref{fig:look}.

We adopt a two-$\Delta$ approach.  First, we generate the lookup table $\mathcal{T}(x||y) = f(x, y)$. For the inputs $x$ and $y$, which need to be concatenated into $x||y$, we treat this as two indexing steps: the first uses $x$ to index the higher $\ell'/2$ bits, and the second uses $y$ to index the lower $\ell'/2$ bits. Based on this idea, during the generation of the lookup table, we group every $2^{\ell'/2}$ small entries into blocks. A global offset $\Delta \in [0, 2^{\ell'/2})$ is generated for the outer blocks, and a shift is applied accordingly. Then, for each block, we apply a same second shift using a local offset $\Delta'$.
Next, we open $\delta = x - \Delta$ and $\delta' = y - \Delta'$, and directly output the result as the $\delta$-th block's $\delta'$-th entry.

\begin{algorithm}[ht]
  \textbf{Input}: lookup table $\mathcal{T} \in \ZZ^{2^{\ell'}}_{2^\ell}$ provided by $P_0$, 2PC additive shared values $\llbracket x \rrbracket^{\ell'/2}$ and $\llbracket y \rrbracket^{\ell'/2}$ held by $P_1$ and $P_2$ \\
  \textbf{Output}: shares of $\mathcal{T}(x||y)$ 
        \begin{algorithmic}[1]
        \State $P_0$ picks random $(\Delta, \Delta') \in \ZZ^2_{2^{\ell'/2}}$;
        \State $P_0$ left-shifts $\mathcal{T}$ by $2^{\ell'/2}\cdot \Delta$ positions;
        \State For $(i,j) \in \ZZ^2_{2^{\ell'/2}}$, $P_0$ moves $\mathcal{T}(i \cdot 2^{\ell'/2} + j)$ to position $i \cdot 2^{\ell'/2} + (j - \Delta' \bmod 2^{\ell'/2})$ of $\mathcal{T}$;
        \State $P_0$ invoke $\llbracket \mathcal{T} \rrbracket^\ell \leftarrow \Pi_\mathsf{share}(\mathcal{T}, P_0)$, $\llbracket \Delta \rrbracket^{\ell'} \leftarrow \Pi_\mathsf{share}(\Delta, P_0)$ and $\llbracket \Delta' \rrbracket^{\ell'} \leftarrow \Pi_\mathsf{share}(\Delta', P_0)$;
        \State $P_1$ and $P_2$ do
        \begin{itemize} 
        \item $\llbracket \delta \rrbracket^{\ell'/2}   = \llbracket x \rrbracket^{\ell'/2} - \llbracket \Delta \rrbracket^{\ell'/2}$;
        \item $\llbracket \delta' \rrbracket^{\ell'/2}   = \llbracket y \rrbracket^{\ell'/2} - \llbracket \Delta' \rrbracket^{\ell'/2}$;
        \item reveal $\delta$ and $\delta'$ to each other;
        \end{itemize}
        \State $P_1$ and $P_2$ set output as $\llbracket \mathcal{T}(\delta \cdot 2^{\ell'/2} + \delta') \rrbracket^{\ell}$;
       \end{algorithmic}
    \caption{Lookup table with separate inputs $\Pi^{\ell'/2, \ell'/2}_{\mathsf{look}}$.}
    \label{alg:separate}
\end{algorithm}

\smallskip
\noindent \textbf{Communication Optimization.}  We observe that in cases where two lookup tables share the same input, the communication cost of table evaluation can be further optimized. Specifically, consider lookup table 1 with inputs $(x, y)$ and lookup table 2 with inputs $(x, y')$, where the input $x$ is reused. Let table 1 use offsets $\Delta^{(1)}$ and $\Delta'^{(1)}$, and table 2 use $\Delta^{(2)}$ and $\Delta'^{(2)}$. By setting $\Delta^{(1)} = \Delta^{(2)}$, i.e., using the same offset when generating the two tables, we only need to open $x-\Delta$ once during the online phase. The revealed value can then be used to query both tables separately. When a large number of lookup tables are involved, this optimization reduces the online communication cost by up to 50\%.

\smallskip
\noindent \textbf{Lookup Table for Share Conversion.} We use lookup tables to perform ring size extension, which expand $\llbracket x \rrbracket^{\ell'}$ to $\llbracket x \rrbracket^{\ell}$, where $\ell > \ell'$. That is, $P_0$ secret shares a lookup table $\mathcal{T}(i) = i$ over the $\ell$-bit ring. Then, using a lookup table evaluation protocol, the input $\llbracket x \rrbracket^{\ell'}$ is converted into a two-party secret share $\llbracket x \rrbracket^{\ell}$ over the $\ell$-bit ring.
 In addition, we employ lookup tables to convert two-party secret shares into replicated secret shares. Specifically, we convert $\llbracket x \rrbracket^{\ell'}$ into $\langle x \rangle^{\ell}$. It works as follows. First, we convert $\llbracket x \rrbracket^{\ell'}$ to $\llbracket x \rrbracket^{\ell}$ using a lookup table. After that, a reshare step is performed to obtain RSS $\langle x \rangle^\ell$, as follows:
\begin{itemize}
\item $P_0$ and $P_1$ jointly generate $\langle x \rangle^\ell_2 $,
\item $P_0$ and $P_2$ jointly generate $\langle x \rangle^\ell_1 $,
\item Then, $P_1$ computes $\delta_1 = \llbracket x \rrbracket^{\ell}_1 - \langle x \rangle^\ell_2 $,
\item $P_2$ computes $\delta_2 = \llbracket x \rrbracket^{\ell}_2 - \langle x \rangle^\ell_1 $,
\item Finally, $P_1$ and $P_2$ open and add the results to obtain $\langle x \rangle^\ell_0 = \delta_1 + \delta_2 $.
\end{itemize}
We denote the above procedure as $\langle x \rangle^\ell \leftarrow \Pi^{\ell', \ell}_{\mathsf{convert}}(\llbracket x \rrbracket^{\ell'})$.

\subsection{Linear Layer Evaluation} \label{sub:newtrc}
Linear layer computation mainly includes Fully Connected layers and Matrix Multiplication. We first discuss the case of Fully Connected layers.

\smallskip
\noindent\textbf{Fully Connection.} The fully connected layer performs matrix operations, which can be reformulated as vector inner products. It takes as input a binary weight vector $W \in \{-1, 1\}^N$, a 4-bit activation vector $x \in \{-8,\ldots,7\}^N$, and three implicit scaling parameters $s_W$, $s_x$ and $s_y$. The computation process is as follows:
$$y = s_w s_x / s_y \cdot \sum^{N-1}_{i=0}W_i \cdot x_i$$
Analyzing the above formula, since each $W_i \cdot x_i$ results in a 4-bit number and the sum of $N$ such 4-bit numbers can be at most a $4 + \log_2 N$-bit number, we need to perform the computation of $\sum W_i \cdot x_i$ over a ring of $4 + \log_2 N$ bits to prevent overflow. Since using RSS for vector inner product computation incurs very low overhead, we employ RSS to perform these inner product operations.
The multiplication by $\frac{s_w s_x}{s_y}$ poses a significant challenge, as this term is a high-precision value. Executing such a computation under MPC requires fixed-point arithmetic and introduces the need for truncation.
Re-examining the calculation, we observe that the purpose of $\frac{s_w s_x}{s_y}$ is to rescale a $4 + \log_2 N$-bit value back down to a 4-bit range. To handle this efficiently, we can scale it by a factor of $2^{12}$, which effectively shifts the original lower 4-bit result to the highest 4 bits. At the same time, $\left\lfloor 2^{12} \cdot \frac{s_w s_x}{s_y} \right\rfloor$ can be treated as an integer, eliminating the need to introduce fixed-point arithmetic.
Then we directly extract the top 4 bits as the final 4-bit result, effectively approximating the rescaling via high-bit truncation.
Considering that in our BERT model, the dimension of the fully connected layer’s inner product is 768, we perform the above computation over a $4 + 12$-bit ring, since $2 ^{12} > 768$. The scaling factor is set to 
$
\left\lfloor 2^{12} \cdot \frac{s_w s_x}{s_y} \right\rfloor
$.
Algorithm~\ref{alg:aby3trc} depicts such procedure. After each multiplication, the output becomes a two-party secret share held by $P_1$ and $P_2$.  For each layer which requires $16$-bit RSS input, if the output of the previous layer is in the form of 4-bit two-party secret sharing, we apply the conversion $\Pi^{4,16}_{\mathsf{convert}}$ to restore it to a 16-bit RSS.

\noindent \underline{Remark:} We observe that under MPC evaluation, clipping is not necessary. The primary purpose of clipping is to prevent extremely large values from affecting the computation. However, in the MPC setting, we find that values exceeding the threshold are effectively truncated and do not impact the evaluation result. 

\smallskip
\noindent\textbf{Matrix Multiplication.} 
Quantized matrix multiplication is similar to a fully connected (FC) layer, as it is also implemented through vector dot products. We can adopt a similar evaluation strategy as used in FC layers.  
Similarly, 
\begin{figure}
\center
\begin{tikzpicture}[x=0.75pt,y=0.75pt,yscale=-0.75,xscale=0.75]

\draw  [fill={rgb, 255:red, 245; green, 166; blue, 35 }  ,fill opacity=0.04 ] (225,152) -- (255,152) -- (255,182) -- (225,182) -- cycle ;
\draw  [fill={rgb, 255:red, 245; green, 166; blue, 35 }  ,fill opacity=0.52 ] (255,152) -- (285,152) -- (285,182) -- (255,182) -- cycle ;
\draw  [fill={rgb, 255:red, 245; green, 166; blue, 35 }  ,fill opacity=0.52 ] (285,152) -- (315,152) -- (315,182) -- (285,182) -- cycle ;
\draw  [fill={rgb, 255:red, 245; green, 166; blue, 35 }  ,fill opacity=0.07 ] (375,152) -- (405,152) -- (405,182) -- (375,182) -- cycle ;
\draw  [fill={rgb, 255:red, 245; green, 166; blue, 35 }  ,fill opacity=0.52 ] (345,152) -- (375,152) -- (375,182) -- (345,182) -- cycle ;
\draw  [fill={rgb, 255:red, 245; green, 166; blue, 35 }  ,fill opacity=0.07 ] (405,152) -- (435,152) -- (435,182) -- (405,182) -- cycle ;
\draw  [fill={rgb, 255:red, 245; green, 166; blue, 35 }  ,fill opacity=0.52 ] (315,152) -- (345,152) -- (345,182) -- (315,182) -- cycle ;
\draw  [fill={rgb, 255:red, 74; green, 144; blue, 226 }  ,fill opacity=0.06 ] (225,100) -- (255,100) -- (255,130) -- (225,130) -- cycle ;
\draw  [fill={rgb, 255:red, 74; green, 144; blue, 226 }  ,fill opacity=0.06 ] (255,100) -- (285,100) -- (285,130) -- (255,130) -- cycle ;
\draw  [fill={rgb, 255:red, 74; green, 144; blue, 226 }  ,fill opacity=0.06 ] (285,100) -- (315,100) -- (315,130) -- (285,130) -- cycle ;
\draw  [fill={rgb, 255:red, 74; green, 144; blue, 226 }  ,fill opacity=0.55 ] (375,100) -- (405,100) -- (405,130) -- (375,130) -- cycle ;
\draw  [fill={rgb, 255:red, 74; green, 144; blue, 226 }  ,fill opacity=0.55 ] (345,100) -- (375,100) -- (375,130) -- (345,130) -- cycle ;
\draw  [fill={rgb, 255:red, 74; green, 144; blue, 226 }  ,fill opacity=0.55 ] (405,100) -- (435,100) -- (435,130) -- (405,130) -- cycle ;
\draw  [fill={rgb, 255:red, 74; green, 144; blue, 226 }  ,fill opacity=0.55 ] (315,100) -- (345,100) -- (345,130) -- (315,130) -- cycle ;
\draw  [fill={rgb, 255:red, 74; green, 144; blue, 226 }  ,fill opacity=0.06 ] (194,100) -- (224,100) -- (224,130) -- (194,130) -- cycle ;
\draw  [fill={rgb, 255:red, 245; green, 166; blue, 35 }  ,fill opacity=0.04 ] (196,152) -- (226,152) -- (226,182) -- (196,182) -- cycle ;
\draw    (159.04,140.96) -- (478.04,140.96) ;
\draw   (339,81.38) -- (345.76,81.38) -- (345.76,70) -- (359.28,70) -- (359.28,81.38) -- (366.04,81.38) -- (352.52,88.96) -- cycle ;
\draw   (323.04,201.58) -- (316.28,201.58) -- (316.28,212.96) -- (302.76,212.96) -- (302.76,201.58) -- (296,201.58) -- (309.52,194) -- cycle ;

\draw (264,159.4) node [anchor=north west][inner sep=0.75pt]  [xscale=0.75,yscale=0.75]  {$1$};
\draw (234,159.4) node [anchor=north west][inner sep=0.75pt]  [xscale=0.75,yscale=0.75]  {$0$};
\draw (294,159.4) node [anchor=north west][inner sep=0.75pt]  [xscale=0.75,yscale=0.75]  {$1$};
\draw (324,159.4) node [anchor=north west][inner sep=0.75pt]  [xscale=0.75,yscale=0.75]  {$0$};
\draw (386,159.4) node [anchor=north west][inner sep=0.75pt]  [xscale=0.75,yscale=0.75]  {$0$};
\draw (356,159.4) node [anchor=north west][inner sep=0.75pt]  [xscale=0.75,yscale=0.75]  {$1$};
\draw (416,159.4) node [anchor=north west][inner sep=0.75pt]  [xscale=0.75,yscale=0.75]  {$1$};
\draw (236,106.4) node [anchor=north west][inner sep=0.75pt]  [xscale=0.75,yscale=0.75]  {$0$};
\draw (266,106.4) node [anchor=north west][inner sep=0.75pt]  [xscale=0.75,yscale=0.75]  {$0$};
\draw (296,106.4) node [anchor=north west][inner sep=0.75pt]  [xscale=0.75,yscale=0.75]  {$0$};
\draw (326,106.4) node [anchor=north west][inner sep=0.75pt]  [xscale=0.75,yscale=0.75]  {$1$};
\draw (356,106.4) node [anchor=north west][inner sep=0.75pt]  [xscale=0.75,yscale=0.75]  {$0$};
\draw (386,106.4) node [anchor=north west][inner sep=0.75pt]  [xscale=0.75,yscale=0.75]  {$0$};
\draw (416,106.4) node [anchor=north west][inner sep=0.75pt]  [xscale=0.75,yscale=0.75]  {$0$};
\draw (205,106.4) node [anchor=north west][inner sep=0.75pt]  [xscale=0.75,yscale=0.75]  {$0$};
\draw (205,159.4) node [anchor=north west][inner sep=0.75pt]  [xscale=0.75,yscale=0.75]  {$0$};
\draw (374,69) node [anchor=north west][inner sep=0.75pt]  [xscale=0.75,yscale=0.75] [align=left] {significant bit};
\draw (329,198) node [anchor=north west][inner sep=0.75pt]  [xscale=0.75,yscale=0.75] [align=left] {significant bit};

\end{tikzpicture}
\caption{Significant bits of division}\label{fig:division}
\end{figure}
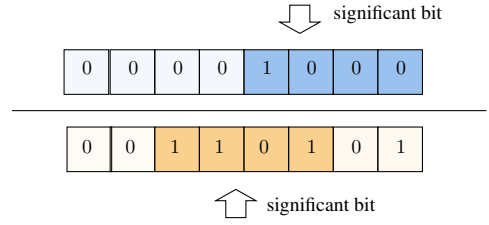

\begin{table*}[ht!]
\centering
\begin{tabular}{cccccccccc} 
\hline
     & \textbf{Bits} & \textbf{MNLI-}m/mm & \textbf{QQP} & \textbf{QNLI} & \textbf{SST-2} & \textbf{STS-B} & \textbf{MRPC} & \textbf{RTE} & \textbf{Avg}  \\ 
\hline
BERT & 32-32         & $84.9/85.5$        & $91.4$       & $92.1$        & $93.2$         & $90.1$         & $86.3$        & $72.2$       & $87.2$        \\ 
\hdashline[1pt/1pt]
Ours & 1-4           & $82.6/82.9$        & $87.2$       & $90.0$        & $91.3$         & $84.5$         & $85.6$        & $58.8$       & $82.9$        \\
\hline
\end{tabular}
\caption{Accuracy benchmark on GLUE benchmark tasks. \textbf{BERT} refers to the original BERT model. \textbf{Bits} denotes weight-activation bit-width. Tasks are evaluated on datasets \textbf{MNLI}, \textbf{QQP}, \textbf{QNLI}, \textbf{SST-2}, \textbf{STS-B}, \textbf{MRPC} and \textbf{RTE}.}\label{tab:glue_results}
\end{table*}

\begin{algorithm}[ht]
  \textbf{Input}: $\langle \cdot \rangle^{16}$ shares of $x_i\in[0,2^{3})\bigcup(2^{16}-2^{3},2^{16})$, $W'_i  = 2^{12} \cdot \frac{s_w s_x}{s_y}  \cdot W_i\in [0,2^{15})\bigcup(2^{16}-2^{15},2^{16})$\\
  \textbf{Output}: shares of $\sum W'_i \cdot x_i$ in $\mathbb{Z}_{2^\ell}$
      \begin{algorithmic}[1]
        \State $P_j$ calculates $y_j = \sum^{N-1}_{i = 0}\langle W'_i\rangle_{j-1} \cdot \langle x_i\rangle_{j+1} + \langle W'_i\rangle_{j+1} \cdot \langle x_i\rangle_{j-1}  + \langle W'_i\rangle_{j+1} \cdot \langle x_i\rangle_{j+1}$.
        \State $P_0$ sends $y_0$ to $P_1$;
        \State $P_1$ set $y_1 = \mathsf{trc}(y_0 + y_1, 4)$;
        \State $P_2$ set $y_2 = \mathsf{trc}(y_2, 4)$;
       \end{algorithmic}
    \caption{Inner product for quantized FC.}
    \label{alg:aby3trc}
\end{algorithm}

\subsection{Nonlinear Layer Evaluation} \label{sub:softmax}

Non-linear computation includes Softmax and ReLU, with the main computational overhead coming from the Softmax operation.

\smallskip
\noindent \textbf{Softmax.} In our framework, the softmax function takes a 4-bit quantized vector as input and outputs a 4-bit quantized softmax result. Similarly, the computation is performed in the dequantized domain. The computation formula is as follows:

$$
\text{softmax}(x_i) = \left\lfloor 2^4 \cdot \frac{e^{x_i \cdot s_x}}{\sum_j e^{x_j \cdot s_x}} \right\rfloor
$$
Here, $x_i \in \{0, 1, \dots, 15\}$ is the 4-bit quantized input, $s_x$ is the input scaling factor used to dequantize the input before exponentiation. The result is rescaled by $2^4$ to fit back into the 4-bit quantized range.

Our secure evaluation scheme is as follows. First, since computing exponentials of $e$ can exceed the computation range, we begin by scaling both the numerator and the denominator~\cite{our2, puma}.
Given that the maximum value of $x$ is $x_o$, we compute $e^{s_x(x_i - x_o)}$, which is always less than or equal to 1. 
Since the softmax output in the model only requires 4-bit precision, we observe that $e^{s_x(x_i - x_o)}$ also only needs to be computed with 4-bit precision. It is easy to see that the numerator will always contain at least one value equal to 1, and the denominator will be greater than 1.
For any input $x_j$ such that $x_j < x_o$, if $e^{s_x(x_j - x_o)} < \frac{1}{16}$, then its contribution to the softmax output will inevitably be zero (after quantization). Therefore, we do not need to retain any data beyond the top 4 bits of $e^{s_x(x_i - x_o)}$.
In particular,  we first employ $\Pi_{\mathsf{max}}$ to calculate the maximum value $x_o$. Then all parties calculate $x_i - x_o$ locally, and  evaluate $e^{s_x(x_i - x_o)}$ using a lookup table $\mathcal{T}(x_i - x_o)$, where input $x_i - x_o$ to the lookup table.
To ensure precision under 4-bit representation, we carefully determine the input bit alignment, as illustrated in Fig.~\ref{fig:division}. First, the lookup table outputs an 8-bit secret share, where the lower 4 bits contain the valid value and the upper 4 bits are zeros. We then perform addition over the 8-bit ring to prevent overflow. As a result, the denominator may carry over into the higher bits. For the numerator, we extract the lower 4 bits; for the denominator, we extract the middle 4 bits—this is because in BERT, the middle 4 bits are sufficient to preserve the high-order information.  Then, we use the previously designed multi-input lookup table protocol $\Pi_{\mathsf{look}}$ to evaluate the division, where the lookup table is defined as $\TTT(x || y) = 2^4 \cdot x / y$.


\smallskip
\noindent \textbf{ReLU.} We follow the work of Lu et al.~\cite{our1} and use the lookup table to evaluate the ReLU function. For a 4-bit input, and considering that the ReLU output will be used as input to the FC layer, we let the lookup table directly output 16-bit additive share, which are then reshared into RSS format.

\smallskip
\noindent \textbf{LayerNorm.} The layer normalization layer requires the computation of the mean and variance, which are expressed as follows:
$$
\mathrm{LN}(x_i) = \gamma \cdot \frac{x_i - \mu}{\sqrt{\sigma^2 + \epsilon}} + \beta
$$
where $\mu$ is the mean of input, and $\sigma^2$ is the variance. We perform layer normalization using 4-bit quantized data over a 16-bit ring. Before feeding the data into this layer, we use a lookup table to convert the 4-bit share $\llbracket x \rrbracket^4$ into 16-bit share $\llbracket x \rrbracket^{16}$. Owing to the homomorphic property of mean computation under quantization, the mean can be directly computed from the quantized values. The procedure is as follows:
\begin{itemize}
\item For quantized vector $(\llbracket x_0 \rrbracket^{16}, \ldots, \llbracket x_{n-1} \rrbracket^{16})$, locally calculate $\llbracket 2^{12} \mu \rrbracket^{16} = \lfloor 2^{12}/n \rfloor \sum^{n-1}_{i = 0} \llbracket x_i \rrbracket^{16}$, which store the result in the highest 4-bit.
\item $P_1$ and $P_2$ set $\llbracket \mu \rrbracket^4_i = \mathsf{trc}(\llbracket 2^{12} \mu \rrbracket^{16}_i, 4)$.
\end{itemize}
Then we convert $\llbracket \mu \rrbracket^4$ to $\langle \mu \rangle^{16}$ by invoking $\Pi^{4, 16}_{\mathsf{convert}}$. Considering that 
$$
\sigma^2 = \frac{1}{n \cdot s_y}\sum_{i=1}^{n} (x_i\cdot s_x - \mu \cdot s_x)^2 =  \frac{s^2_x}{s_y\cdot n} \sum_{i=1}^{n} (x_i - \mu)^2
$$
where $s_x$ is the quantized scaler for $x$, and $s_y$ is the quantized scaler for variance,  we perform the multiplications over $\langle \cdot \rangle$ to compute $\sum_{i=1}^{n} (x_i - \mu)^2$. Similar to the mean computation, we multiply the result by $\lfloor 2^{12}\frac{s^2_x}{s_y\cdot n} \rfloor$ and apply truncation to obtain a 4-bit representation. Consequently, we employ the lookup table with two 4-bit inputs to evaluate the final division.


\begin{table}[t]
\centering
\small 
\caption{Performance comparison with CrypTen and Sigma. “\#x” denotes a CPU with x threads.}
\label{tab:benchmark1}
\begin{tabular}{lcccccc}
\toprule
CrypTen & Sigma & Sigma & \multicolumn{3}{c}{Ours} \\
\cmidrule(lr){4-6}
GPU & \#4 & GPU & \#4 & \#20 & \#96 \\
\midrule
21551 & 12311.4 & 4667.9 & 1315.4 & 1165.1 & 969.5 \\
\bottomrule
\end{tabular}
\end{table}

\begin{table}[ht]
\centering
\small
\caption{Performance comparison with Lu et al. under the WAN setting.}\label{tab:benchmark2}
\begin{tabular}{rrrrr}
\toprule
\makecell{Sequence\\Length} & Lu.\ et al& 20 threads & 96 threads & Speedup \\
\midrule
8  & 8135.61 & 1239.29 & 1037.55 & $\times$7.84 \\
16 & 12143.00 &  1591.61 & 1485.85 & $\times$8.17 \\
32 & 16764.15 &  2320.04 & 2143.16 & $\times$7.82 \\
\bottomrule
\end{tabular}
\end{table}

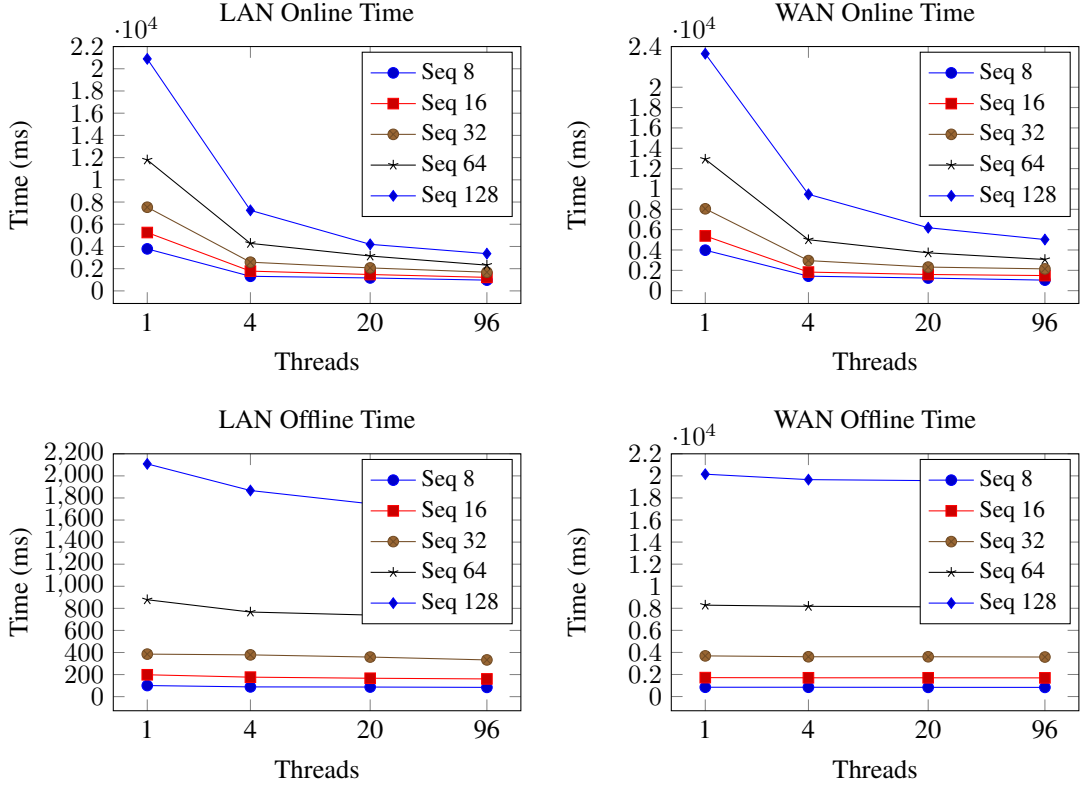
\begin{figure*}[t!] 
  \centering 
  \begin{tikzpicture}
\begin{groupplot}[
  group style={group size=2 by 2, horizontal sep=2cm, vertical sep=2cm},
  width=7cm, height=5cm,
  xlabel={Threads},
  xmode=log,
  ylabel={Time (ms)},
  legend style={font=\small, cells={anchor=west}},
  xtick={1,4,20,96},
  xticklabels={1,4,20,96}
]

\nextgroupplot[
  title={LAN Online Time},
  ytick={0,2000,4000,6000,8000,10000,12000,14000,16000,18000,20000,22000},
  ymax=22000
]
\addplot coordinates {(1,3782.88) (4,1315.43) (20,1165.08) (96,969.505)};
\addlegendentry{Seq 8}
\addplot coordinates {(1,5255.59) (4,1780.14) (20,1474.76) (96,1218.96)};
\addlegendentry{Seq 16}
\addplot coordinates {(1,7533.58) (4,2584.55) (20,2068.89) (96,1681.78)};
\addlegendentry{Seq 32}
\addplot coordinates {(1,11791.6) (4,4278.5) (20,3140.12) (96,2319.06)};
\addlegendentry{Seq 64}
\addplot coordinates {(1,20890.2) (4,7253.94) (20,4192.54) (96,3352.78)};
\addlegendentry{Seq 128}

\nextgroupplot[
  title={WAN Online Time},
  ytick={0,2000,4000,6000,8000,10000,12000,14000,16000,18000,20000,24000},
  ymax=24000
]
\addplot coordinates {(1,3982.48) (4,1425.14) (20,1239.29) (96,1037.55)};
\addlegendentry{Seq 8}
\addplot coordinates {(1,5384.8) (4,1828.3) (20,1591.6) (96,1485.85)};
\addlegendentry{Seq 16}
\addplot coordinates {(1,8049.21) (4,2956.49) (20,2320.04) (96,2143.16)};
\addlegendentry{Seq 32}
\addplot coordinates {(1,12924.8) (4,5010.69) (20,3719.95) (96,3065.43)};
\addlegendentry{Seq 64}
\addplot coordinates {(1,23297.4) (4,9472.84) (20,6189.21) (96,5027.59)};
\addlegendentry{Seq 128}

\nextgroupplot[
  title={LAN Offline Time},
  ytick={0,200,400,600,800,1000,1200,1400,1600,1800,2000,2200},
  ymax=2200
]
\addplot coordinates {(1,100.537) (4,88.35) (20,87.015) (96,83.989)};
\addlegendentry{Seq 8}
\addplot coordinates {(1,198.193) (4,176.599) (20,166.709) (96,160.643)};
\addlegendentry{Seq 16}
\addplot coordinates {(1,385.275) (4,378.456) (20,358.696) (96,332.764)};
\addlegendentry{Seq 32}
\addplot coordinates {(1,877.951) (4,766.491) (20,739.115) (96,714.494)};
\addlegendentry{Seq 64}
\addplot coordinates {(1,2108.51) (4,1867.43) (20,1747.39) (96,1609.93)};
\addlegendentry{Seq 128}

\nextgroupplot[
  title={WAN Offline Time},
  ytick={0,2000,4000,6000,8000,10000,12000,14000,16000,18000,20000,22000},
  ymax=22000
]
\addplot coordinates {(1,843.106) (4,841.897) (20,833.93) (96,830.06)};
\addlegendentry{Seq 8}
\addplot coordinates {(1,1715.22) (4,1700.51) (20,1698.98) (96,1692.26)};
\addlegendentry{Seq 16}
\addplot coordinates {(1,3685.43) (4,3606.07) (20,3605.98) (96,3581.51)};
\addlegendentry{Seq 32}
\addplot coordinates {(1,8281.84) (4,8176.07) (20,8132.21) (96,8120.49)};
\addlegendentry{Seq 64}
\addplot coordinates {(1,20152.8) (4,19659.6) (20,19572.4) (96,18697.3)};
\addlegendentry{Seq 128}

\end{groupplot}
\end{tikzpicture}
  \caption{The latency of our privacy-preserving BERT for generating the output of a single token under different network settings, numbers of threads, and input sequence lengths.}\label{fig:oursper}
\end{figure*}

\begin{table}[ht]
\centering
\caption{Communication cost comparison with CrypTen and Sigma (MB)}\label{tab:communication_cost}
\begin{tabular}{rcccc}
\toprule
Tokens & Our Online & Our Offline & CrypTen & Sigma \\
\midrule
8  & 4.43   & 29.20  & 3921  & 43.28 \\
16 & 8.87 & 59.34  & 8342  & 89.24 \\
32 & 17.80 & 122.46 & 21114 & 189.17 \\
64 & 35.83 & 260.01 & 63731 & 421.09 \\
\bottomrule
\end{tabular}
\end{table}

\section{Performance Evaluation} \label{sec:evaluate}
In this section, we benchmark our new technique with regard to performance improvement and the effect of accuracy on model inference. 
%
We then compare the performance of our quantized BERT model inference with CrypTen~\cite{CrypTen}, Sigma~\cite{sigma} and Lu. et al.~\cite{our1}. 
We use three cloud server nodes to simulate three parties, each node with the following configuration: two Intel(R) Xeon(R) E5-2690 v4 @ 2.60GHz CPUs, 64 GiB memory. They are equipped with  Ubuntu 16.04.7. We also simulate two different network environments: LAN and WAN corresponded to 5Gbps/100Mbps bandwidth and 0.2ms/40ms round trip latency, respectively. We divide the execution of our framework into two phases: the offline phase and the online phase. Since the random lookup tables can be generated before the inputs are available, we let $P_0$ generate and distribute these tables during the offline phase.

\noindent\textbf{Accuricy of Our Scheme} \label{sub:keybit} We evaluated the accuracy of our privacy-preserving quantized BERT inference on the GLUE benchmark, as shown in Table~\ref{tab:glue_results}. “BERT” represents the accuracy of the original full-precision model (32–32 bits), while “Ours” corresponds to our quantized privacy-preserving approach with 1–4 bit precision. Across all GLUE tasks, our method achieves competitive performance with only a moderate drop in average accuracy from 87.2 to 82.9.

\noindent\textbf{Communication Cost Comparison.} Table.~\ref{tab:communication_cost} presents the communication cost comparison between our approach and existing privacy-preserving inference frameworks (CrypTen and Sigma) across different sequence lengths. Our method demonstrates significant communication efficiency improvements. 
Compared to CrypTen, our method achieves a $885\thicksim 1778\times$ reduction in online communication cost ($116\thicksim 215\times$ overall when including the offline phase).
Against Sigma, our online communication cost shows substantial improvements as $9.8 \thicksim 11.8 \times$ reduction in online communication cost and $1.3 \thicksim 1.4 \times$ overall cost.

\smallskip
\noindent\textbf{Performance Benchmark.}  \label{sub:unittest}
Fig.~\ref{fig:oursper} evaluates the latency of our privacy-preserving BERT for generating the output of a single token under different network settings, numbers of threads, and input sequence lengths. The results include both the offline phase  and the online phase. It can be observed that increasing the number of threads from 1 to 20 significantly improves the performance of the online phase. With 20 threads and an input length of 8 tokens, the online evaluation of our quantized BERT requires about 1 second. Even with 128 tokens, the evaluation under LAN conditions finishes within 4 seconds.

\noindent\textbf{Performance Comparison.} \label{sub:e2etest} 
We conduct experiments on end-to-end BERT inference, comparing it to CrypTen~\cite{CrypTen} and Sigma~\cite{sigma} on different CPU threads under LAN setting. 
The results of the experiments are presented in Table.~\ref{tab:benchmark1}. 
With 4 threads, Sigma takes $12$ s, whereas our method only requires 1.3 s, achieving a $9.36\times$ speedup over Sigma. We did not implement a GPU-based version; however, even without GPU acceleration, our 96-thread CPU implementation is still $22\times$ faster than CrypTen and $4.8\times$ faster than Sigma on GPU.

In Table.~\ref{tab:benchmark2}, we compare the performance of our approach with that of Lu et al. under the WAN setting for quantized privacy-preserving inference. We evaluate input lengths ranging from 8 to 32 tokens, while the results of Lu et al. are reported with 96 threads. With 96 threads, our approach is $7$--$8\times$ faster than Lu et al., and even with only 20 threads, our method still outperforms their 96-thread implementation by $6.5$--$7.5\times$.

\bibliography{myref}
%

%


\end{document}